\documentclass[11pt,a4paper]{article}
\usepackage[hyperref]{emnlp2020}
\usepackage{times}
\usepackage{latexsym}
\usepackage{hyperref}
\usepackage{color}
\usepackage{xcolor}
\usepackage{todonotes}
\usepackage{caption}
\usepackage{pgfplots}
\usepackage{booktabs}
\pgfplotsset{width=7cm,compat=1.8}
\usepackage[acronym]{glossaries}
\usepackage{algorithm} 
\usepackage{algpseudocode} 
\usepackage{amsmath}
\usepackage{xspace}
\usepackage{enumitem}
\usepackage{tcolorbox}
\usepackage{subcaption}

\DeclareMathOperator*{\argminA}{arg\,min}

\usepackage{microtype}

\usepackage[belowskip=-10pt,aboveskip=9pt]{caption}
\def\aclpaperid{***} 

\newacronym{asr}{ASR}{Automatic Speech Recognition}
\newacronym{nlu}{NLU}{Natural Language Understanding}
\newacronym{slu}{SLU}{Spoken Language Understanding}
\newacronym{wer}{WER}{Word Error Rate}
\newacronym{tp}{TP}{True Positive}
\newacronym{fp}{FP}{False Positive}
\newacronym{fn}{FN}{False Negative}
\newacronym{etoe}{E2E}{End-to-end}
\newacronym{eeslu}{E2E-SLU}{End-to-end SLU}
\newacronym{lm}{LM}{Language Model}
\newacronym{crf}{CRF}{Conditional Random Field}
\newacronym{lstm}{LSTM}{Long Short-Term Memory}
\newacronym{bilstm}{BiLSTM}{Bidirectional Long Short-Term Memory}
\newacronym{tts}{TTS}{Text-To-Speech}

\newcommand{\metricname}{SLU-F1}
\newcommand{\werm}{Word-F1}
\newcommand{\levm}{Char-F1}
\newcommand{\datasetacr}{SLURP}
\newcommand{\datasetname}{Spoken Language Understanding Resource Package}
\newcommand{\etoe}{E2E}
\newcommand{\slu}{SLU}
\newcommand{\asr}{ASR}
\newcommand{\nlu}{NLU}
\newcommand{\wer}{WER}
\newcommand{\tp}{TPs}
\newcommand{\fp}{FPs}
\newcommand{\fn}{FNs}
\newcommand{\datasetwholek}{72k}

\newcommand{\scenarion}{18}
\newcommand{\actionn}{46}
\newcommand{\entityn}{55}
\newcommand{\hermit}{HerMiT}
\newcommand{\sfid}{SF-ID}
\newcommand{\gasr}{Google-ASR}
\newcommand{\masr}{Multi-ASR}

\def\VRdel#1{\bgroup\markoverwith{\textcolor{magenta}{\rule[0.5ex]{2pt}{1pt}}}\ULon{#1}}

\aclfinalcopy

\title{\datasetacr: A \datasetname}
  
  \author{Emanuele Bastianelli$^{\dagger}$$^*$\thanks{$^*$Authors contributed equally.}, 
  Andrea Vanzo$^{\dagger}$$^*$, 
  Pawel Swietojanski$^{\ddagger}$$^*$ and Verena Rieser$^\dagger$ \\
  $^\dagger$The Interaction Lab, MACS, Heriot-Watt University, Edinburgh, UK \\
  $^\ddagger$Faculty of Engineering, University of New South Wales, Sydney, Australia \\
  {\tt \{e.bastianelli, a.vanzo, v.t.rieser\}@hw.ac.uk} \\
  {\tt p.swietojanski@unsw.edu.au} \\
  }

\date{}

\begin{document}

\maketitle
\begin{abstract}
Spoken Language Understanding infers semantic meaning directly from audio data, and thus promises to reduce error propagation and misunderstandings in end-user applications. However, publicly available SLU resources are limited. In this paper, we release \datasetacr, a new SLU package  containing the following: (1) A new challenging dataset in English spanning \scenarion{} domains, which is substantially bigger and linguistically more diverse than existing datasets; (2) Competitive baselines based on state-of-the-art NLU and ASR systems; (3)  A new transparent metric for entity labelling which enables a detailed error analysis for identifying potential areas of improvement. \datasetacr{} is available at \url{https://github.com/pswietojanski/slurp}
\end{abstract}
\section{Introduction}

Traditionally, Spoken Language Understanding (\slu) 
uses a pipeline transcribing audio into text using Automatic Speech Recognition (\asr), which 
is then mapped
into a semantic structure via Natural Language Understanding (\nlu). 
However, this modular approach is prone to error propagation from noisy \asr{} transcriptions, and \asr{} in turn is not able to disambiguate based on semantic information. 
End-to-end (\etoe) approaches 
on the other hand, can benefit from joint modelling.
One of the main bottlenecks for building \etoe-\slu{} systems, however, is the lack of large and diverse 
 datasets
of audio inputs paired with corresponding semantic structures. 
Publicly available datasets to date are limited in terms of lexical and semantic richness \cite{lugosh19:Interspeech}, number of vocalizations \cite{Coucke18:Snips}, domain coverage \cite{Hemphill90:ATIS, Dahl:94ATIS} and semantic contexts \cite{Godfrey92:switchboard, Jurafsky1997:DASwitchboard}. 
In this paper, we present the \datasetname{} (\datasetacr), a publicly available multi-domain dataset for \etoe-\slu{}, which is substantially bigger and 
more diverse than existing SLU datasets. \datasetacr{} is a collection of \texttildelow\datasetwholek{} audio recordings of single turn user interactions with a home assistant, annotated with three levels of semantics:  Scenario, Action and Entities, as in Fig.\ \ref{fig:data}, 
including over \scenarion{} different scenarios, with \actionn{} defined actions and \entityn{} different entity types as listed on \url{https://github.com/pswietojanski/slurp}.\footnote{Note that Action \& Entities are also referred to as `Intent'. Entities consist of `Tags' and  `Fillers', aka. `Slots' and 'Values'.}

\begin{figure}
\begin{tcolorbox}
\footnotesize
\begin{description}[noitemsep]
\item[User:] ``{\em Make a calendar entry for brunch on Saturday morning with Aaronson.}"
\item[Scenario:] Calendar
\item[Action:] Create\_entry
\item[Entity tags and lexical fillers:] [{\tt event\_name}: brunch], [{\tt date}: Saturday], [{\tt timeofday}: morning], [{\tt person}: Aaronson]
\end{description}
\end{tcolorbox}
\caption{Example annotation from \datasetacr{} dataset.} 
    \label{fig:data}
\end{figure}

In order to further support \slu{} development, we propose \metricname, a new metric for entity prediction, 
 which is specifically designed to assess  error propagation in structured \etoe-\slu{} tasks. 
 This metric has 3 main advantages over the commonly used accuracy/F1 metric, 
 aimed at supporting \slu{} developers: First, it computes a distribution rather than a single score. This distribution is (1) inspectable and interpretable by system developers, and (2) can  be converted into a confidence score which can be used in the system logic (akin to previously available ASR confidence scores). Finally, the distribution reflects errors introduced by ASR and their impact on NLU and thus (3) gives an indication of the scope of improvement that can be gained by \etoe{} approaches.
Using this metric, we evaluate 4 baseline systems that represent competitive pipeline approaches, i.e.\ 2 state-of-the-art \nlu{} systems and 2 \asr{} engines. We conduct a detailed error analysis of cases where \etoe{} could have made a difference, i.e.\ error propagation and semantic disambiguation. 

\section{Related Work}
\label{sec:related}

The first corpora containing both audio and semantic annotation reach as far back as the Air Travel Information System (ATIS) corpus \cite{Hemphill90:ATIS} and the Switchboard-DAMSL Labeling Project \cite{Jurafsky1997:DASwitchboard}.
However, it was not until recently when the first \etoe{} approaches to \slu{} were introduced \cite{Serdyuk18:towards,Haghani2018:FromAT}. Since then, one of the main research questions is how to overcome data sparsity by e.g. using transfer learning  \cite{Schuster19:facebook, tomashenko:hal-02307811}, or pre-training \cite{lugosh19:Interspeech}.
Here, we present a new corpus, SLURP, which is considerably bigger than previously available corpora. 
In particular, we directly compare our dataset to the two biggest \etoe-\slu{} datasets for the English language: 
The Snips benchmark \cite{Coucke18:Snips} and the 
Fluent Speech Command (FSC) corpus \cite{lugosh19:Interspeech}.
With respect to these resources, 
\datasetacr ~contains \texttildelow6 times more sentences than Snips,
\texttildelow2.5 times more audio examples than FSC, while covering 9 times more domains and being on average 10 times lexically richer than both FSC and Snips, see 
 Section \ref{subsec:comparison}. SLURP represents the first E2E-SLU corpus of this size for the English language. The only existing comparable project is represented by the CASTLU dataset \cite{10.1145/3340555.3356098} for Chinese Mandarin.


\section{\datasetacr{} data}
\label{sec:dataset}

\subsection{Data Collection}

\datasetacr{} was collected for
developing an in-home personal robot assistant~\cite{miksik2020building}.
First, we collected textual data by prompting Mechanical Turk (AMT) workers to formulate commands towards the robot, using 200 pre-defined prompts such as ``How would you ask for the time/ set an alarm/ play your favourite music?" etc. We carefully designed the prompts to avoid lexical priming and 
thus increase linguistic variability of the collected data. 
This data has been manually annotated 
at scenario, action and entity level, and released as a text-only NLU benchmark \cite{liu19:benchmarking}. 
The textual data also serves as gold standard transcriptions for the audio data.

The audio data was collected in acoustic conditions matched to a typical home or office environment. 
We asked 100+ participants to read out the collected prompts on 
a tablet and 
to provide demographic background information, see Table \ref{tab:dem}. 
Speech was captured at distance with a microphone array, but some users were also equipped with a close-talking headset microphone (though, distant and close-talk channels are not synchronised at the sample level). 
Most recording sessions lasted 1 hour and were split into 4 parts. In each part, the 
 technician changed position of the microphone array in the collection place. Users were encouraged to vary their location in the room from utterance to utterance (seating, standing or walking), and for some utterances not to speak directly to the mic array in order to resemble realistic conditions. These parameters are not logged with the dataset, however, they do pose increased challenges for \asr{} \cite{Marino2011AnAO}.

\begin{table}[!h]
    \centering
    \footnotesize
    \begin{tabular}{c|c|c|c|c}
\toprule
Female & Male & Native & Non-Native & Unk.\\
\midrule
37.3\% & 32.2\% & 25.5\% & 44\% & 30.5\% \\
\bottomrule
\end{tabular}
    \caption{Participants' demographic statistics.}
    \label{tab:dem}
\end{table}

\subsection{Audio Data Processing}
\begin{figure}
\centering
\includegraphics[width=1.0\columnwidth]{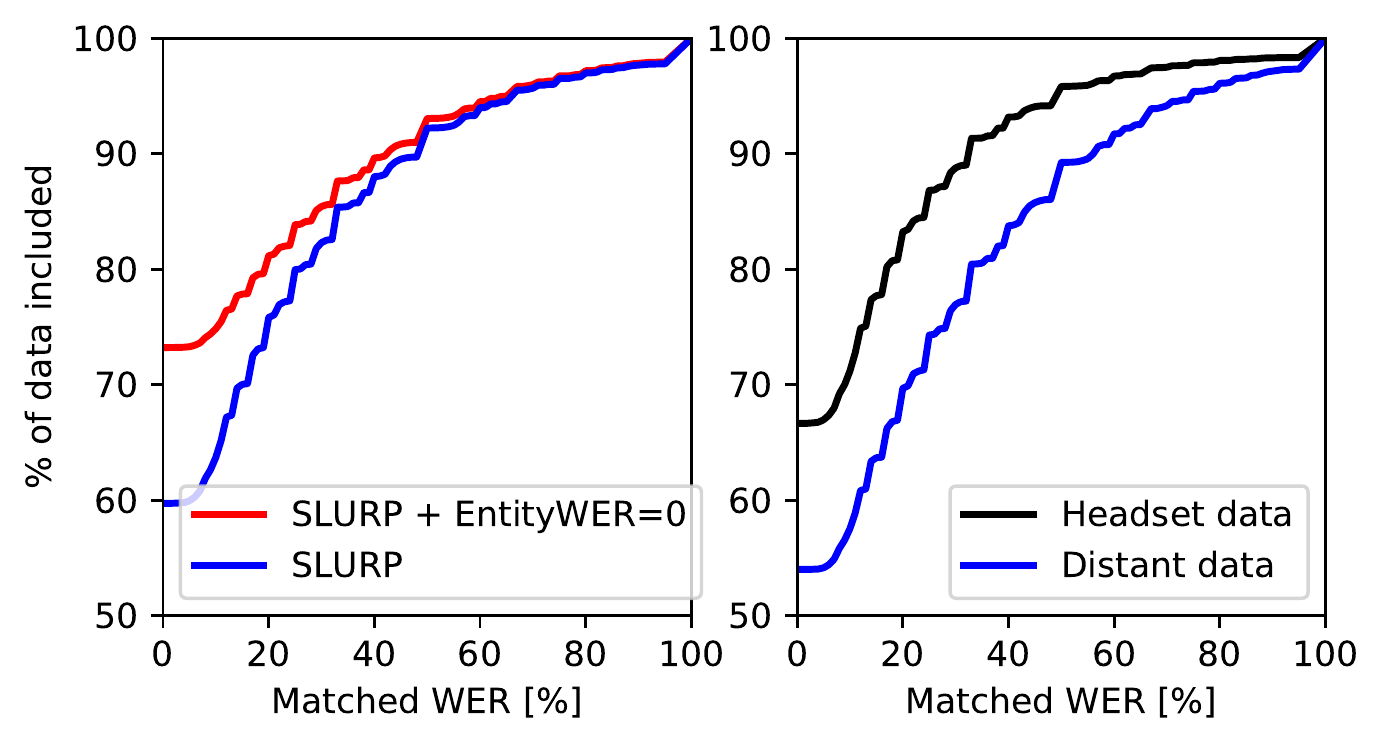}
\caption{Amounts of data in \datasetacr{} matching given \wer{} levels.} 
\label{fig:audio_sel}
\end{figure}
For quality control of the audio data, we automatically verified i) whether the participant uttered the right / complete \slu{} query as prompted and ii) if the files were appropriately end-pointed.  
We used 
the transcriptions of
two \asr{} systems 
(referred to as \masr~and \gasr, see Sec~\ref{sec:asr}). 
These systems were not estimated from \datasetacr{} acoustic data, thus remain unbiased and do not reinforce potential errors. First, we removed all data that failed to force-align to transcripts using \masr. Then for the remainder we derived the \slu{} related confidences based on the matched Word-Error Rate (\wer) between textual prompts and the obtained \asr{} hypotheses (calculated for both utterance and entity fillers), as well as cross-mic validation between close and distant microphones, see Figure~\ref{fig:audio_sel} (Right). Note that the higher matched \wer{} does not necessarily imply the file lacks the expected content, as simply the file could be more challenging to automatically recognise. At the same time, from \slu{} perspective, one does not necessarily need grammatically correct utterances, as long as the they carry the information necessary to understand and execute the query.
Figure~\ref{fig:audio_sel} (Left) shows 
that for nearly 60\% of the data at least one ASR system achieved a perfect score (\wer=0), and this increases to \texttildelow73\% after including utterances with imperfect sentence error rates but correct entity fillers (EntityWER=0). 
After filtering, \datasetacr{} comprises \texttildelow58 hours of acoustic material.  
See Table~\ref{tab:audio} for detailed statistics.

In addition, we provide \datasetacr-synth following  \cite{arxiv.org/abs/1910.09463}, where we replace filtered or missing recordings with synthetic vocalisations from Google's Text-to-Speech 
system\footnote{\url{https://cloud.google.com/text-to-speech}} using 34 different synthetic English voices. 




\subsection{Linguistic Analysis and Comparison}
\label{subsec:comparison}
In this Section, we compare \datasetacr{} 
with the most recent publicly available \etoe-\slu{} datasets: The Fluent Speech Command (FSC) corpus \cite{lugosh19:Interspeech} and the Snips benchmark \cite{Coucke18:Snips}, which are also set in the smart-home domain.
 Snips 
 covers 10 domains. However, only 2 domains
 have been vocalised, 
 resulting in \texttildelow6K audio files. FSC, on the other hand, is considerably bigger than Snips in terms of audio recordings, including \texttildelow30k vocalisations. However, the provided semantics only cover a small subset of actions with no more than two fixed entity types as arguments.
In the following, we compare these dataset along 
four dimensions in order to get a first estimate of \datasetacr's level of complexity. 

\begin{table}[t]
    \centering
    \footnotesize
    \begin{tabular}{lrrrr}
\hline
         &    \multicolumn{1}{c}{FSC} &   \multicolumn{1}{c}{Snips} &    \multicolumn{1}{c}{\datasetacr} &   \multicolumn{1}{c}{\datasetacr}\\
         &    &    &   & -synth\\
\hline
 Speakers      &    97    &   69    &  {\bf 177}  
 &     34    \\
 Audio files         & 30,043    & 5,886    & \textbf{72,277}    &     69,253\\
 \; -- Close range &    30,043    &  2,943    & \textbf{34,603}    &      --\\
 \; -- Far range     &    --    &   2,943    & \textbf{37,674}    &     --\\
 Audio/Sentence  &   \textbf{121.14} &    2.02 &     4.21 &        3.87\\
 Duration [hrs] & 19 & 5.5 & \textbf{58} & 43.5  \\
 Avg. length [s] & 2.3 & \textbf{3.4} & 2.9 & 2.3 \\
\hline
\end{tabular}
    \caption{Audio file statistics.} 
    \label{tab:audio}
\end{table}

\noindent{\bf Audio analysis:} Table \ref{tab:audio} 
summarises 
 the audio data for each dataset. 
 Audio files are differentiated in \textit{close} and \textit{far} range microphone. 
 As shown, \datasetacr{} has \texttildelow 1.8$\times$ more speakers, more than double the audio files than the biggest dataset FSC, however FSC has an higher
 audio-per-sentence ratio. Demographic statistics are reported in Table~\ref{tab:dem}. 

\begin{table*}[!ht]
    \centering
    \footnotesize
\begin{tabular}{lrrrrrrrr}
\hline
        & \multicolumn{2}{c}{FSC}      & \multicolumn{2}{c}{Snips}       & \multicolumn{2}{c}{\datasetacr}       & \multicolumn{2}{c}{\datasetacr-synt}   \\
    &   \multicolumn{1}{c}{\textsc{Lex}} & \multicolumn{1}{c}{\textsc{Delex}} &   \multicolumn{1}{c}{\textsc{Lex}} & \multicolumn{1}{c}{\textsc{Delex}} &   \multicolumn{1}{c}{\textsc{Lex}} & \multicolumn{1}{c}{\textsc{Delex}} &   \multicolumn{1}{c}{\textsc{Lex}} & \multicolumn{1}{c}{\textsc{Delex}} \\
\hline
Sentences &248 &190 &2,912 &1,437 &\textbf{17,181} &\textbf{15,433} &19,711 & 16,707 \\
Average Sentence length  & 4.49 & -- & \textbf{7.48} & -- & 6.93 & -- & 7.27 & -- \\
Distinct Tokens &96 &89 &2,182 &271 &\textbf{6,467} &\textbf{3,774} &5,974 & 3,553 \\
Distinct Tokens occurring once &31 &36 &1825 &120 &\textbf{3,007} &\textbf{1,778} &2,799 & 1,676 \\
Distinct Lemmas &102 &92 &2193 &250 &\textbf{5,501} &\textbf{3,080} &5,119 & 2,920 \\
Distinct Bigrams &218 &182 &4,004 &1,355 &\textbf{32,303} &\textbf{21,724} &28,988 & 20,308 \\
Distinct Bigrams occurring once &97 &97 &3,066 &698 &\textbf{21,997} &\textbf{14,095} &19,360 & 12,637 \\
Distinct Trigrams &250 &198 &5,703 &2,408 &\textbf{50,422} &\textbf{37,417} &45,631 & 35,548 \\
Distinct Trigrams occurring once &131 &119 &4,499 &1,543 &\textbf{40,184} &\textbf{28,393} &34,856 & 25,553 \\
Lexical Sophistication (LS2) &0.35 &0.31 &\textbf{0.87} &0.41 &0.79 &\textbf{0.69} &0.79 & 0.68 \\
Corrected Verb Sophistication (CVS1) &0.42 &0.38 &0.72 &0.59 &\textbf{5.17} &\textbf{3.54} &4.58 & 3.20 \\
Mean segmental TTR (MSTTR) & 0.71 & 0.82 & 0.78 & 0.86 &\textbf{0.92} &\textbf{0.96} & 0.93 & 0.96 \\
\hline
\end{tabular}
    \caption{Analysis of Lexical diversity and sophistication. 
    }
    \label{tab:lexical}
\end{table*}

\noindent{\bf Lexical analysis:} 
Table \ref{tab:lexical} provides an overview of different measures of lexical richness and diversity, following \cite{novikova-etal-2017-e2e}, using both 
lexicalised (\textsc{Lex}) and delexicalised (\textsc{Delex}) versions of the datasets (delexicalisation is performed by replacing each entity span with the entity label).  
 Note that delexicalisation has a more severe effect on FSC and Snips, which indicates that most of their lexical richness and diversity stems from entity names. 
On average \datasetacr{} has \texttildelow100$\times$ more 
tokens, lemmas, bigrams and trigrams 
than FSC, and \texttildelow10$\times$ more than Snips.
In addition, we compute the following
lexicographic measures 
using the Lexical Complexity Analyser \cite{lu2012:lex}. \textit{Lexical Sophistication} (LS2) \cite{laufer94:lex} is defined as $T_s/T$, with $T_s$ being the number of sophisticated types of (unique) words\footnote{Sophisticated words are considered words not in the 2000 more frequent words in English language.} and $T$ being the number of types of words in a dataset. The \textit{Corrected Verb Sophistication} (CSV1) \cite{wolfe1998:second} is evaluated as $T_{svb}/\sqrt{2N_{vb}}$, with $T_{svb}$ the number of types of sophisticated verbs and $N_{vb}$ the total number of verbs in a dataset. The \textit{Mean Segmental Text-to-Token Ratio} (MSTTR) \cite{wendell44:lex} is the average Text-to-Token Ratio (TTR -- $T/N$) over all the segments of 10\footnote{Standard size of a segment for written text is 50, but we are here considering short utterances, so we lowered this number to 10.} words, with $N$ the number of words in a dataset. The MSTTR is used to capture the variation of classes of words. Again, \datasetacr{} shows higher levels of lexical sophistication and richness than the other datasets, especially in the delexicalised case. Note that lexicalised version of Snips contains many names of artists and bands in the music scenario, which contributes to enlarge the set of sophisticated words $T_s$.
The only measure where \datasetacr{} doesn't outperform the other datasets is
average sentence length. 
\datasetacr{} contains, among others,  shorter interactions, such short acknowledgements, elliptic questions and atomic commands, 
whereas Snips is mostly composed of commands of similar length, often including multiword named entities. 

\noindent{\bf Syntactic analysis:} Next, we use the D-Level Analyser \cite{lu09:synt} to evaluate the syntactic complexity of user utterances 
 according to the revised D-Level scale \cite{covington06:dlevel}, where higher levels correspond to more complex, deeper syntactic structures, e.g. 0-1 levels include simple sentences, while higher levels presents embedded structures, subordinating conjunction, etc. Figure \ref{plt:syntax} shows the percentages on the D-Level scale for each dataset. Overall, all the datasets present a majority of Level 0 and 1 sentences. This can be explained with the nature of the application domain, i.e. a smart-home assistant. 
 FSC contains mostly Level 0 sentences (\texttildelow89\%), with some (\texttildelow9\%) Level 4 ones. 
 89\% of Snips sentences fall into Level 0 and 1, against only 74\% of \datasetacr. The remaining 11\% of Snips are mostly Level 4 sentences, while \datasetacr{} appears more mixed, with even a \texttildelow5\% of Level 7 sentences.
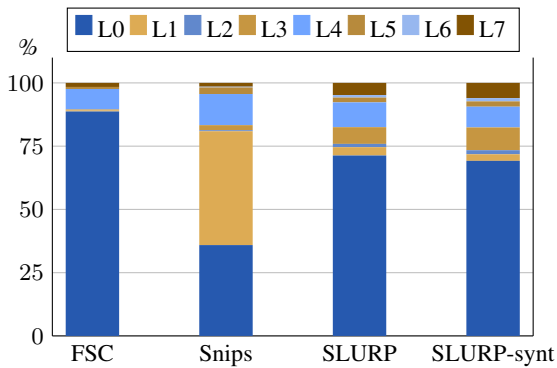
\begin{figure}
\centering
\footnotesize
\begin{tikzpicture}
\definecolor{clr0}{RGB}{38,89,175}
\definecolor{clr1}{RGB}{221,171,84}
\definecolor{clr2}{RGB}{97,136,204}
\definecolor{clr3}{RGB}{198,150,65}
\definecolor{clr4}{RGB}{112,164,255}
\definecolor{clr5}{RGB}{183,138,60}
\definecolor{clr6}{RGB}{151,183,239}
\definecolor{clr7}{RGB}{130,83,3}
\begin{axis}[
    ybar stacked,
    bar width=20pt,
    width=225pt,
    height=150pt,
    ymin=0,
    enlargelimits=0.1,
    legend style={at={(0.03,1.1)}, anchor=west},
    ylabel={\%},
    y label style={at={(axis description cs:-0.05,1.1)}, rotate=90,anchor=south},
    symbolic x coords={FSC, Snips, \datasetacr, \datasetacr-synt},
    xtick=data,
    enlarge y limits={upper=0},
    legend columns=-1,
    ymajorgrids = true,
    ytick={0,25,50,75,100},
    xtick style={draw=none},
	ytick style={draw=none},
    axis x line*=bottom,
    ]
\addplot+[draw=none, ybar, clr0] plot coordinates {(FSC, 88.7) (Snips, 35.9)  (\datasetacr, 71.3) (\datasetacr-synt, 69.2)};
\addplot+[draw=none, ybar, clr1] plot coordinates {(FSC, 0.4) (Snips, 45.0) 
(\datasetacr, 3.3)  (\datasetacr-synt, 2.6)};
\addplot+[draw=none, ybar, clr2] plot coordinates {(FSC, 0.0) (Snips, 0.4)
(\datasetacr, 1.3) (\datasetacr-synt, 1.6)};
\addplot+[draw=none, ybar, clr3] plot coordinates {(FSC, 0.4) (Snips, 2.0)
(\datasetacr, 6.6)  (\datasetacr-synt, 9.0)};
\addplot+[draw=none, ybar, clr4] plot coordinates {(FSC, 8.1) (Snips, 12.3) 
(\datasetacr, 9.8)  (\datasetacr-synt, 8.2)};
\addplot+[draw=none, ybar, clr5] plot coordinates {(FSC, 0.8) (Snips, 2.6)
(\datasetacr, 1.9)  (\datasetacr-synt, 2.1)};
\addplot+[draw=none, ybar, clr6] plot coordinates {(FSC, 0.0) (Snips, 0.3)
(\datasetacr, 1.0)  (\datasetacr-synt, 1.3)};
\addplot+[draw=none, ybar, clr7] plot coordinates {(FSC, 1.6) (Snips, 1.5)
(\datasetacr, 4.8)  (\datasetacr-synt, 6)};
\legend{\strut L0, \strut L1, \strut L2, \strut L3, \strut L4, \strut L5, \strut L6, \strut L7}
\end{axis}
\end{tikzpicture}
\caption{Syntactic complexity on D-Level scale, where higher levels correspond to more complex, deeper syntactic structures.}
\label{plt:syntax}
\end{figure}

\noindent{\bf Semantic Analysis:} Finally, we compare the datasets according to their semantic content. \datasetacr{} is annotated with three layers of semantics, namely \textit{scenarios}, 
\textit{actions}  and \textit{entities}, 
where each sentence is annotated with one scenario and one action,  see Fig.\ \ref{fig:data}, similar to 
annotations used in \cite{budzianowski2018:woz, Schuster19:facebook}. FSC and Snips contain actions and entities as well, although they do not explicitly annotate the scenarios, however these can be deducted from the dataset file structure. 
The results in Table \ref{tab:semantics} show 
that \datasetacr's semantic coverage is 9 times wider than other datasets in terms of scenarios, and \texttildelow6.5 times in terms of actions, where  a higher number of scenarios results in a higher number of actions. 
FSC has the highest entity/sentence ratio, though it only has 16 unique entities. Snips appears to be the dataset with highest Unique Entities/Total Entities ratio, \texttildelow50\%, against \texttildelow33\% of \datasetacr. Again, this is due to the frequent use of proper names. 

\begin{table}[]
    \centering
    \footnotesize
    \begin{tabular}{lrrrr}
\hline
            & \multicolumn{1}{c}{FSC}   & \multicolumn{1}{c}{Snips}   &   \multicolumn{1}{c}{\datasetacr} &   \multicolumn{1}{c}{\datasetacr}\\
            & & & & -synt\\
\hline
 Scenarios  & 2       & 2     &      \textbf{18} &     18       \\
 Actions    & 6       & 7      &      \textbf{46} &     54       \\
 Entities     & 2        & 4       &      \textbf{56} &  56          \\
 Tot. Entities        & 334      & 2,870    &   \textbf{16,792} &     14,623    \\
 Entity/Sentence & \textbf{1.35} & 0.98  & 0.97 &  0.65 \\
 Unique Entities & 16       & 1,348    &    \textbf{5,613} &    4619      \\
\hline
\end{tabular}
    \caption{Semantic analysis of the number of scenarios, actions and entity types, the total number of annotated entities, 
and the number of unique entities, i.e. entities whose lexical filler appears only once.}
    \label{tab:semantics}
\end{table}

\section{\datasetacr{} Metrics}
\label{sec:measure}

The standard metric for evaluating \etoe-\slu{} is accuracy, 
which is defined as ``the accuracy of all slots for an utterance taken together -- that is, if the predicted intent differs from the true intent in even one slot, the prediction is deemed incorrect'' \cite{lugosh19:Interspeech}. 
However, this notion of accuracy is problematic when it comes to evaluating entities, as it does not account for the interplay between semantic mislabelling and textual misalignment. Nor does it differentiate between entity label and lexical filler, as  in Fig.\ \ref{fig:slu_exp}, where lexical filler is defined as span over tokens in the original sentence.\footnote{In traditional \nlu{} systems this is identified with pairs of start-end tokens or chars, or token index spans.}

\begin{figure}
\begin{tcolorbox}
\footnotesize
\begin{description}[noitemsep]
\item[Gold:] [{\tt event\_name}: brunch], [{\tt date}: Saturday], [{\tt timeofday}: morning], [{\tt person}: Aaronson]
\item[SLU:] [{\tt event\_name}: brunch], [{\tt date}: Saturday], [{\tt date}: morning], [{\tt person}: Aron's son]
\end{description}
\end{tcolorbox}
\caption{Continued example from Figure~\ref{fig:data}: Errors in \slu{} entity tagging.}
    \label{fig:slu_exp}
\end{figure}
Formally, given a sentence $s$, let $\mathcal{E}$ and $\mathcal{\hat{E}}$ be the set of gold and predicted entities, respectively. Each $e_i = \langle l_i, f_i\rangle \in \mathcal{E}$ is a tuple where $l_i \in \mathcal{L}$ is the label drawn from the list of available entity labels $\mathcal{L}$, while $f_i = [t_m, \ldots, t_n]$ is the lexical filler, defined as a span of consecutive tokens of $s$ such that $1 \leq m \leq n \leq |s|$. Similarly, predicted entities are of the form $\hat{e}_k = \langle \hat{l}_k, \hat{f}_k\rangle \in \mathcal{\hat{E}}$.
In 
span-based metrics, two entities $e_1$ and $e_2$ are identical ($e_1 =:= e_2$) when both labels and lexical fillers are the same ($l_1 = l_2 \land f_1 = f_2$). A match is thus found only whenever the gold and predicted entities are identical, i.e. $e_i =:= \hat{e}_k$. 
This evaluation method holds in \nlu{} because entities are tagged over the same textual sequence. When evaluating \etoe-\slu, where entities are identified out of a wave form, this strict coupling with the token sequence may no longer apply. 
Note that pipeline systems for \slu{} are affected as well since they operate over \asr{} transcribed sentences, which can consistently differ from the original gold transcription. 

To account for this mismatch, we propose \metricname, a new metric which does not  overly penalise misalignments caused by \asr{} errors. In addition, 
it is able to capture the quality of transcriptions and entity tagging errors at the same time in a single metric. As such, this  metric allows to directly compare \etoe{}  and pipeline systems. 
In particular, \metricname{} combines span-based F1 evaluation with a text-based distance measure \textit{dist}, e.g. \wer. The equality property $=:=$ is  relaxed by 
allowing gold and predicted entities ($e_i$ and $\hat{e}_k$) to match ($e_i =:= \hat{e}_k$) when the corresponding labels are identical ($l_i = \hat{l}_k$), even when the fillers are not identical. In this case we increment the True Positives (\tp) by 1. To account for lexical distance/ mismatch, we compute the \textit{dist} between gold and predicted fillers ($dist(f_i, \hat{f}_k)$), and increment the False Positives (\fp) and False Negatives (\fn) of this amount, as in Algorithm \ref{lst:werf1}. In the case of a predicted entity label  matching with more than one gold entity, e.g.\ when two or more entities with the same label are present, we opt for a non-conservative approach, selecting the gold annotation minimising the \textit{dist} as a candidate. The assumption is that the pair of entities is most likely referring to the same text span. 
We use two distance functions to capture different aspects of possible transcription mistakes:  \wer{} 
(\werm) 
and the normalised Levenshtein distance on character level 
(\levm). \wer{} is a strict token-level metric, which outputs errors/null matches whenever a mismatching or misalignment of tokens is observed. 
The character-based Levenshtein distance, on the other hand, offers the opposite perspective. 
By computing character-based similarities, it is much less susceptible to small variations of input strings, and thus better accounting for local transcription errors which do not affect \nlu{} tagging.
For example, \werm ~will penalise small morphological differences e.g. singular vs. plural as in \textit{pizza} vs. \textit{pizzas}, which are often seen in transcriptions. This  over-penalises \nlu{} outputs, e.g. the tagging of \textit{pizzas} may be semantically correct. \levm ~on the other hand does not over-penalise \nlu, but it also may provide a positive score when two fillers have similar characters, but are semantically and phonetically unrelated.
In other words, \werm{} shows the influence of \asr{} on \nlu, whereas \levm{} gives an indication of \nlu{} performance despite transcription noise.
These \textit{dist}-F1 metrics ($dist=$ Word or Char) metric are similar to the fuzzy matching mechanism proposed in \cite{48919}. They fundamentally differ for the adopted string matching schema: any \textit{dist}-F1 considers string ordering to score string similarity, while the fuzzy mechanism is instead order invariant.

Consider the illustrative entity tagging example in Figure \ref{fig:slu_exp}. 
Here, \textit{Aaronson} has been wrongly transcribed into \textit{Aron's son}, and \textit{morning} has been wrongly tagged with \texttt{date}. A $dist-$F1 will score the predicted entities as follows:
both [\texttt{event\_name}: \textit{brunch}] and [\texttt{date}: \textit{Saturday}] contribute with a $+1$ to the \tp{}, since both label and filler correspond to gold information. The wrong label associated with \textit{morning} increases the \fp{} of $1$, although it is correctly transcribed. It follows that the entity \texttt{timeofday} is not predicted, increasing the \fn{} of 1. Finally, [\texttt{person}: \textit{Aron's son}] is correctly labelled, but its filler is partially wrong. It thus contributes to the \tp{} by 1, but \fp{} and \fn{} are both incremented by $dist(\textit{Aaronson}, \textit{Aron's son})$. 
\begin{algorithm}
    \footnotesize
	\caption{$dist$-F1 for a sentence $s$}
	\textbf{Input} $\mathcal{E}$, $\mathcal{\hat{E}}$,\\
    \hspace*{\algorithmicindent} \hspace{0.2cm} $TP, FP, FN \leftarrow 0$ \\
    \hspace*{\algorithmicindent} \hspace{0.2cm} $\mathcal{L}_s\leftarrow$ set of gold entity labels in $s$\\
    \hspace*{\algorithmicindent} \hspace{0.2cm} $dist \leftarrow$ a text-based distance metric\\
    \textbf{Output}: $TP, FP, FN$
	\begin{algorithmic}[1]
		\For {\textbf{each} $\hat{e} \in \mathcal{\hat{E}}$}
		    \If {$\hat{e}.label \in \mathcal{L}_s$}
		        \State $\mathcal{P}_l\leftarrow \{(e, \hat{e}) \; | \; \forall e \in \mathcal{E} . \; e.label = \hat{e}.label \}$ 
		        \If {$\mathcal{P}_l.size > 0$}
		            \State $(e, \hat{e})\leftarrow \argminA_{(e, \hat{e}) \in \mathcal{P}_l} dist(e, \hat{e})$
		            \State $TP$ += 1
		            \State $FP$ += $dist(e.filler, \hat{e}.filler)$
		            \State $FN$ += $dist(e.filler, \hat{e}.filler)$
		            \State $\mathcal{E}.remove(e), \; \mathcal{\hat{E}}.remove(\hat{e})$
		        \Else
		            \State $FP$ += 1, $\mathcal{\hat{E}}.remove(\hat{e})$
		        \EndIf
		   \Else
		        \State $FP$ += 1, $\mathcal{\hat{E}}.remove(\hat{e})$
		   \EndIf
		\EndFor
		\For {$e \in \mathcal{E}$}
		    \State $FN$ += 1, $\mathcal{E}.remove(e)$
		\EndFor
	\end{algorithmic} 
	\label{lst:werf1} 
\end{algorithm}

Finally, we combine \werm{} and \levm{} in a single number
\metricname, which evaluates the final performance over the sum of the confusion matrices obtained with \werm{} and \levm.\footnote{The official script for analysis and evaluation will be released with SLURP at \url{https://github.com/pswietojanski/slurp}.} 

\section{Experiments}
\label{sec:results}
We now establish the performance of different baseline systems on the \datasetacr{} corpus. As demonstrated in Section 3.1, \datasetacr{} is  linguistically more diverse than previous datasets, and therefore more challenging for \slu.
We first provide an evaluation of two \asr{} baselines to show the complexity of the acoustic dimension.
We then evaluate the semantic dimension, by testing the corpus against state-of-the-art \nlu{} systems.
We finally combine \asr{} and \nlu, implementing several \slu{} pipelines.


Note that so far, the direct comparison of \etoe-\slu{} with pipeline approaches are mainly limited to baselines developed on the same dataset, e.g. a multistage neural model in which the two stages that correspond to
\asr{} and \nlu{} are trained independently, but using the same training data \cite{desot:asru2019,Haghani2018:FromAT}. We follow a different approach, which, as we argue, is closer to the real-life application scenario: We use competitive \asr{} systems  and state-of-the-art \nlu{} systems.

\subsection{Acoustic evaluation} \label{sec:asr}
We run the analysis of the \datasetacr{}  acoustic complexity 
by testing 2 different \asr{} systems:
 In-domain ASR trained on \datasetacr{} data, and \masr{}, which leverages a large amount of out-of-domain data.
  Both are built with the Kaldi \asr{} toolkit~\cite{povey2011kaldi}. 
%
\masr ~is a large-scale system estimated 
from publicly available acoustic data pooled together -- Acoustic data including, among others, LibriSpeech~\cite{panayotov2015librispeech}, Switchboard~\cite{Godfrey92:switchboard}, Fisher~\cite{cieri2004fisher}, CommonVoice~\cite{ardila2019common}, AMI~\cite{carletta2007unleashing} and ICSI~\cite{janin2003icsi},
which is further augmented to increase environmental robustness following~\cite{ko2017study}\footnote{System build while third author was with Emotech LTD.}. 
In total, a time-delay neural network acoustic model~\cite{peddinti2015time} is trained on 24,000 hours of augmented audio material with lattice-free maximum mutual information objective~\cite{povey2016purely}. For decoding, we use a tri-gram Language Model (LM) that is an interpolation of an in-domain LM estimated from 60k voice-command sentences\footnote{This includes SLURP-Train and additional 50k sentences that has been collected, but not annotated for \nlu{} purposes.} and a background LM estimated from Fisher transcripts. As shown in the first block of Table~\ref{tab:wers_generic}, \masr{} offers a competitive performance on this data when compared to the off-the-shelf Google-ASR.\footnote{\url{https://cloud.google.com/speech-to-text/} tested on 20/05/2020 using the \texttt{command\_and\_search} model. Note, that these systems are not directly comparable as \masr~benefits from speaker adaptation, and an in-domain LM data.}

\datasetacr-ASR shares the overall pipeline with \masr, except the acoustic model is estimated from the 40 hours of \datasetacr{} training data (83 hours when pooled with \datasetacr-Synth) 
and bootstrapped from forced-alignments obtained with Gaussian mixture model build for \masr. Results for this scenario are reported in the second block of Table~\ref{tab:wers_generic}, where adding synthetic data shows 1.6\% improvement. For comparison, estimating acoustic models from synthetic data alone (no augmentations) results in 98\% \wer{} on Test partition. 

Finally, we perform supervised acoustic domain adaptation~\cite{bell2020adaptation} of \masr~with \datasetacr-Train by a method proposed in~\cite{swietojanski2016learning}, which achieves the best performance 
 by around 1\% absolute on Test.

In sum, the large out-of-domain \masr{} system performs better than the systems trained on in-domain \datasetacr{} data. Best results are achieved by using a pre-training approach, i.e.\ \masr{} adapted to \datasetacr. This shows that, despite \datasetacr's absolute size, 
the acoustic data is still too scarce to fully account for its lexical richness and noise conditions. As such, \datasetacr{} is a challenging dataset for \asr{} as well as for \slu.


\begin{table}[t]
	\centering
	\small
	\begin{tabular}{l|c|c}
	\toprule
	&  Dev & Test \\ 
	\midrule
    \gasr  & 24.0 & 24.7 \\
    \masr  & 16.7 & 17.3 \\
    \midrule
    \datasetacr{}-ASR (Train) & 23.7  & 23.8 \\
    \datasetacr{}-ASR (Train + Synth) & 22.4 & 22.2 \\ 
    \midrule
    \masr~+ Adapt w/ \datasetacr  &  {\bf 16.3} & {\bf 16.2} \\
    \bottomrule
	\end{tabular}
	\caption{\datasetacr{} \wer{} for different \asr{} systems.}
	\label{tab:wers_generic}
\end{table}

\subsection{Semantic evaluation}

\noindent{\bf System Descriptions:} We evaluate \datasetacr{} against two state-of-the-art \nlu{} models:
\hermit{} \cite{vanzo:2019b} and \sfid{} \cite{e-etal-2019-novel}. Both systems achieved state-of-the-art results on the NLU Benchmark \cite{liu19:benchmarking} and on ATIS/Snips respectively.
\hermit's architecture is a hierarchy of self-attention mechanisms and Bidirectional Long Short-Term Memory (BiLSTM) encoders followed by Conditoinal Random Field (CRF) tagging layers. Its multi-layered structure resembles a  top-down approach of 
 Scenario, Action and, Entity prediction, where each task benefits from the information encoded by the previous stages, 
 e.g.\ Entity detection can benefit from sentence-level encodings. 

\sfid's architecture is also based on attention, using a BiLSTM encoder and CRF tagger. The model defines two subnets that communicate through a reinforce vector. 
In order to compare with \hermit's top-down approach, we choose the opposite Entity-first propagation direction for \sfid, i.e.\ the entity detection task is executed first and its encodings are used to feed the Intent detection task.
Note that while \hermit{} uses a multi-layered annotation scheme (Scenario and Action), \sfid{} can only handle a single layer of annotation. To this end, we generate another combined semantic layer, Scen\_Act, to feed \sfid{} with a label composed by the concatenation Scenario and Action.


\begin{table}[t]
	\centering
	\footnotesize
	\begin{tabular}{l|ccc}
	\toprule
	& Train & Dev & Test\\
	\midrule
	Sentences & 11514 & 2033 & 2974 \\
    Audio files & 50628 & 8690 & 13078 \\
    Tot. Entities & 11367 & 2022 & 2823 \\
    Entity/Sentence & 0.98 & 0.99 & 0.95 \\
    Total duration [hours]  & 40.2 & 6.9 & 10.3 \\
    \bottomrule
	\end{tabular}
	\caption{Data distribution of train, dev and test sets.}
	\label{tab:split}
\end{table}


\begin{table}[t]
	\centering
	\footnotesize
	\begin{tabular}{l|ccc}
	\toprule
	& \textbf{Scenario} & \textbf{Action} & \textbf{Scen\_Act}\\
	\midrule
	Gold/\hermit & {\bf 90.15} & {\bf 86.99} & {\bf 84.84} \\
	Gold/\sfid & 86.48 & 83.69 & 82.25 \\
	\midrule
    Multi/\hermit & 83.73 & 79.70 & 76.68 \\
    Multi/\sfid & 81.90 & 77.72 & 75.87 \\
    Google/\hermit & 81.68 & 76.58 & 73.41\\
    Google/\sfid & 78.87 & 74.31 & 72.06\\
    \midrule 
    SLURP/\hermit & 82.31 & 78.07 & 74.62 \\
    {\scriptsize Multi-SLURP/\hermit} & {\bf 85.69} & {\bf 81.42} & {\bf 78.33} \\
    \bottomrule
   	\end{tabular}
	\caption{System accuracy of Scenario and Action.}
	\label{tab:intents}
\end{table}



\noindent{\bf Scenario and Action Prediction:}
We split SLURP in train, development and test 
as in Table \ref{tab:split}.
We first evaluate accuracy for Scenario, Action and a combination of the two.
Table \ref{tab:intents} summarises the results, where 
the top two rows 
 are upper bounds based on gold transcriptions.  
Note that even for the gold transcriptions, both NLU systems perform substantially below their state-of-the-art results on the NLU benchmark (\hermit=$87.55$) and Snips respectively (\sfid = $97.43$). This further demonstrates the complexity of \datasetacr, which also makes it a challenging test bed for future research not only for SLU, but also NLU.
When moving on to ASR transcribed data, the results in the middle of Table \ref{tab:intents}
 show the \masr{} system in combination with \hermit{} achieves top performance for all 3 tasks. Finally, the 3rd block reports  \hermit{} with ASR from in-domain SLURP audio data (also see Table \ref{tab:wers_generic}).
 The results show that our best performing system, \hermit{} with
 \masr~+ Adapt w/ \datasetacr, is only \texttildelow5\% below the gold standard despite 16\% WER.
We hypothesise that this is due to robust Scenario and Action encodings, which we will further examine in our error analysis in Section \ref{sec:error}.

\begin{table}[t]
	\centering
	\footnotesize
	\begin{tabular}{l|p{0.8cm}p{0.8cm}p{0.8cm}p{0.8cm}p{0.8cm}}
	\toprule
	& \textbf{Word-F1} & \textbf{\levm} & \textbf{\metricname} & \textbf{F1} \\
	\midrule
	Gold/\hermit & -- & --  & -- & {\bf 78.19} \\
	Gold/\sfid & -- & -- & -- & 69.87 \\
	\midrule
    Multi/\hermit & 67.78 & 71.38 & 69.53 & 62.69 \\
    Multi/\sfid & 65.82 & 68.92 & 67.33  & 60.15 \\
    Google/\hermit & 64.01 & 68.12 & 66.00 & 58.00\\
    Google/\sfid & 62.73 & 65.37 & 64.02 & 56.54\\
    \midrule 
     SLURP/\hermit & 65.48 & 68.56 & 66.99 & 59.79 \\
    {\scriptsize Multi-SLURP/\hermit} & \bf 69.34 &  \bf 72.39 & \bf 70.84 &  \bf 64.16 \\
    \bottomrule
	\end{tabular}
	\caption{System performance on entity prediction} 
	\label{tab:entities}
\end{table}

\noindent{\bf Entity Prediction:}
We now analyse the results for entity prediction in more detail using our proposed metric \metricname.
The results in Table \ref{tab:entities} confirm that \hermit{} is the stronger NLU system on gold-transcribed data
 and  outperforms the other system combinations for SLU
 in combination with \masr.
Again, these results suggest that the top-down information flow of \hermit{} (i.e.\ first decoding Scenario, then Action and lastly Entity in a sequence) is better suited for this complex dataset, which we will further demonstrate in the following.

\section{Error Analysis}\label{sec:error}


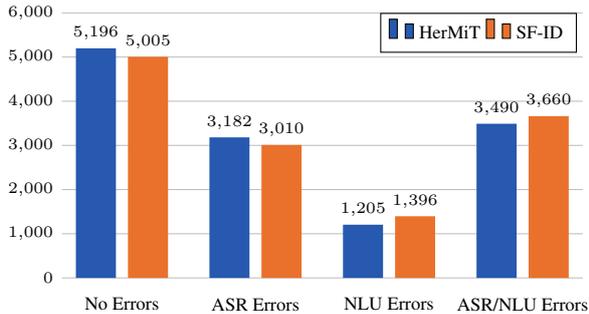
\begin{figure}
\centering
\footnotesize
\begin{tikzpicture}
\definecolor{clr0}{RGB}{38,89,175}
\definecolor{clr1}{RGB}{234,113,43}
\begin{axis}[
	bar width=15pt,
	width=240pt,
	height=145pt,
	ymin=0,
	ymax=6000,
	axis line style={draw=none},
	ytick={0,1000,2000,3000,4000,5000,6000},
	xtick style={draw=none},
	ytick style={draw=none},
	ybar=4.5pt,
    legend style={at={(0.8,1.0)}, anchor=north, font=\scriptsize},
    symbolic x coords={No Errors, ASR Errors, NLU Errors, ASR/NLU Errors},
    nodes near coords,
    xtick=data,
    every node near coord/.append style={font=\tiny},
    legend columns=-1,
    every y tick label/.append style={font=\tiny},
    every x tick label/.append style={font=\scriptsize},
    ymajorgrids = true,
    enlarge x limits=0.15,
    nodes near coords align={vertical},
    ]
\addplot[draw=none, fill=clr0] plot coordinates {(No Errors, 5196) (ASR Errors, 3182) (NLU Errors, 1205) (ASR/NLU Errors, 3490)};
\addplot[draw=none, fill=clr1] plot coordinates {(No Errors, 5005) (ASR Errors, 3010) (NLU Errors, 1396) (ASR/NLU Errors, 3660)};
\legend{HerMiT, SF-ID}
\end{axis}
\end{tikzpicture}
\caption{Error propagation: \textit{No Errors} refer to the number of predicted entities that match the gold transcriptions perfectly. %
\textit{\asr{} Errors} count the number of predictions where \asr{} outputs an unmatched candidate 
but the \nlu{} system is nevertheless able to recover the correct entities from the transcriptions. 
\textit{NLU Errors} count sentences where transcriptions are correct, but entities do not match. 
 \textit{ASR/NLU Errors} count the sentences where both \asr{} and \nlu{} errors are present.}
\label{plt:error}
\end{figure}

\begin{figure*}[t]
\centering
  \begin{subfigure}[b]{0.45\textwidth}
    \includegraphics[width=\textwidth]{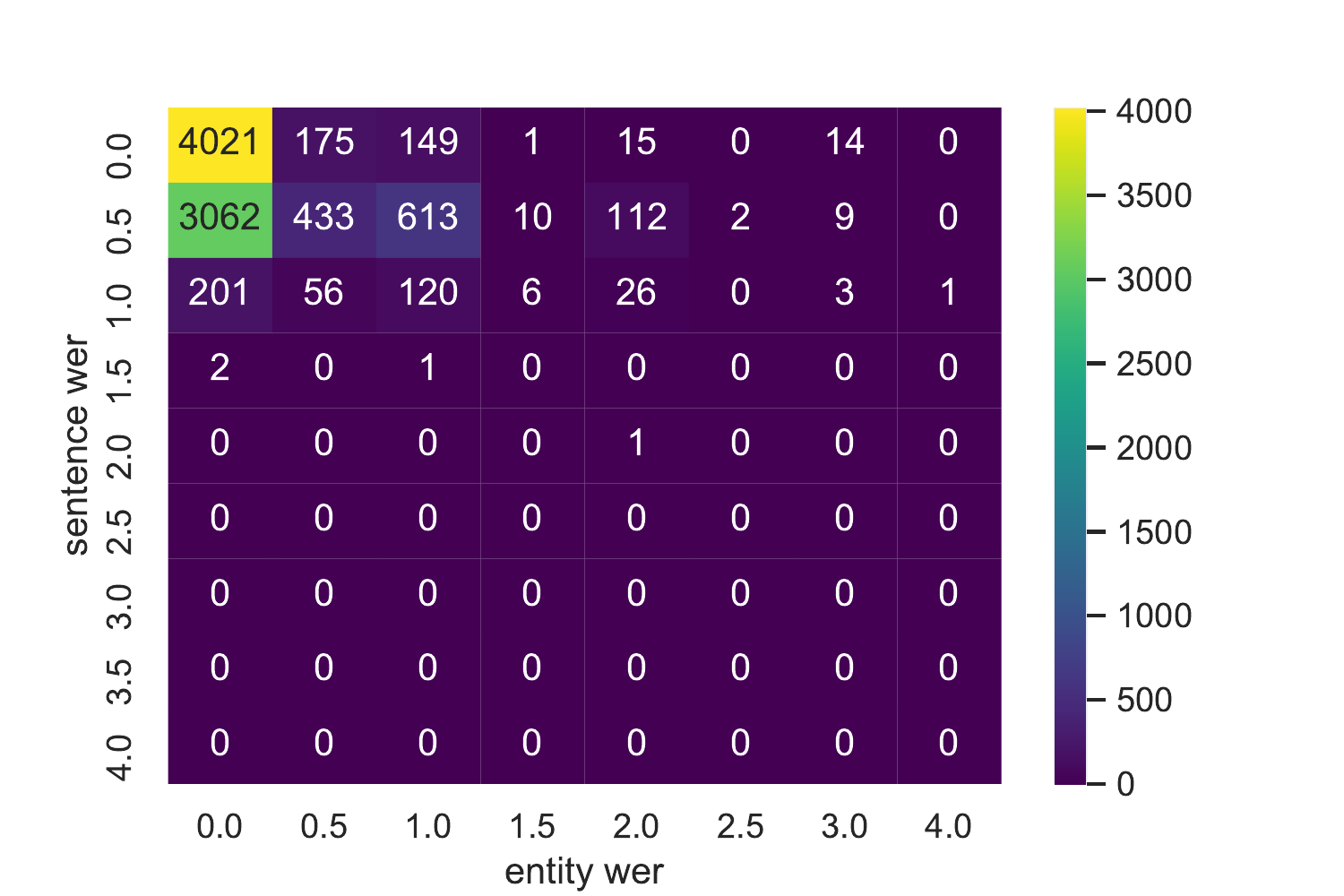}
    \caption{\label{plt:werf1}}
  \end{subfigure}
  \begin{subfigure}[b]{0.45\textwidth}
    \includegraphics[width=\textwidth]{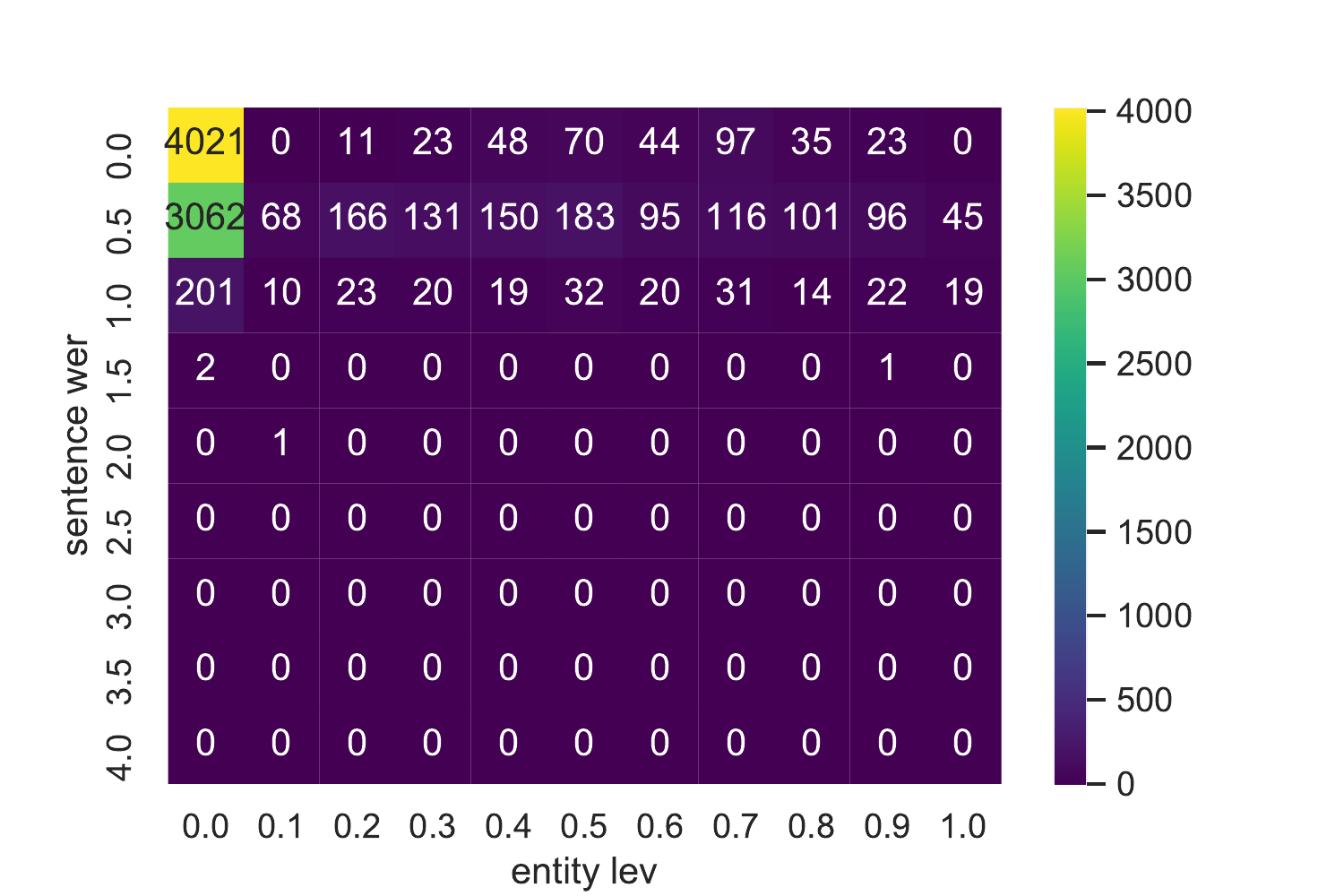}
    \caption{\label{plt:levf1}}
  \end{subfigure}
      \caption{Correlation between sentence-level WER (intervals of 0.5) and entity-level (a) WER values (intervals of 0.5), (b) normalised character-based Levenshtein values (intervals of 0.1).\label{plt:correlation}} 
\end{figure*}

\subsection{Analysis of Error Propagation for different NLU Approaches} 
We further describe the types of errors produced by \hermit{} and \sfid{} for Entity Prediction on noisy ASR data, as shown in Figure \ref{plt:error}.
Overall, \hermit{} has lower error rates for all but ASR errors. Nevertheless, it is able to recover the correct entities from the transcriptions. 
These results indicate that \hermit{}, using a top-down decoding approach -- going from the more general Scenario to the more specific Action and Entity Prediction, is more robust to noise propagation than the bottom-up \sfid{} system.

\subsection{Expressiveness of the \metricname{} Metric}
The results in Table \ref{tab:entities} show that our proposed metrics \werm{} and \levm{} both produce the same ordering as F1. 
However, 
a Pearson's correlation between 
\werm{} and \levm{} shows that the two metrics are  only weakly correlated ($\rho= 0.2$, $p\ll0.0001)$, which confirms that they are indeed measuring two different aspects despite producing the same final ordering.
In addition to an overall performance score, the metrics give us a distribution of value ranges, which
can give us insight on system behaviour. 
Figure \ref{plt:correlation} shows distributions of
entity-level $dist$ value ranges 
over the \wer{} of the sentence for 
our top performing system \hermit/\masr.
For entity-WER (Figure \ref{plt:werf1}), the distribution shows high density of entities falling between sentence-WER$=[0,1]$ and entity-WER$=[0,1]$. When analysing sentences with correct transcriptions, i.e. sentence-WER=$0$, we find only \nlu{} errors, due to span misalignments. When sentence-WER $>0$, most of the entities are scored with a values either in $(0,0.5]$, or in $(0.5,1]$. In the first case, we find \nlu{} mistakes caused by shortening entity spans, e.g. ``football'' instead of ``football match''. 
The second range includes span  shortening and extensions, e.g. ``Saturday morning'' instead of ``Saturday'', as well as many 
mis-transcribed entities, e.g.\ due to either morphological errors (singular vs. plural), or transcription errors. 

The distribution for entity-level normalised Levenshtein is less spiked, as shown in Fig.\ \ref{plt:levf1}. As for WER, all the entries with sentence-WER=0 and entity-Lev$>$0 correspond to correctly labelled entities, whose span has been shortened or extended. 
Entities assigned with character-based Lev values falling between $(0,0.2]$ mostly contain negligible ASR errors, such as morphological errors, 
compound merging or explosion, or general transcription mistakes, e.g.\ \textit{Sara} vs. \textit{Sarah}. Entities with Lev$=(0.2,0.5]$ comprise both \asr{} errors, as well as including minor \nlu{} errors such as shortened or extended entity spans. 
When entity-Lev$=(0.5,0.8]$, we find mostly \nlu{} errors due to wrong span tagging. Finally, two types of \nlu{} errors fall in the range $(0.8,1.0]$: Either span errors with a substantial mismatch in length with gold annotations, or more severe \asr{} errors. 
 



\section{Discussion}
\label{sec:discussion}

\datasetacr{} is not only bigger, but also a magnitude more challenging than previous datasets. The purpose of this new data release is not to provide yet another benchmark dataset, but to provide a use-case inspired new challenge, which is currently beyond the capabilities of SOTA E2E approaches (due to scalability, lack of data efficiency, etc.). 

We have tested several SOTA \etoe-\slu{} systems on \datasetacr, including \cite{lugosh19:Interspeech} which produces SOTA results on the FSC corpus. However, re-training these models on this more complex domain did not converge or result in meaningful outputs. Note that these models were developed to solve much easier tasks (e.g.\ a single domain). Developing an appropriate model architecture is left for future work. For this reason, in this work we focus on benchmarking existing approaches.

We show that SOTA modular approaches are able to provide a strong baseline for this challenging data, which has yet to be met by SOTA E2E systems. We also argue that our modular baseline is closer to how real-world applications build SLU systems, nevertheless often overlooked when testing E2E systems. As such, we consider our SOTA modular baseline a major novel contribution.

\section{Conclusion}
\label{sec:conclusion}

In this paper, we present \datasetacr, a new resource package for \slu. 
First, we present a novel dataset, which is substantially bigger than other publicly available resources. We show that this dataset is also more challenging by first conducting a linguistic analysis, and then demonstrating the reduced performance of state-of-the-art \asr{} and \nlu{} systems.
Second, we propose the new \metricname{} metric for evaluating entity prediction in \slu{} tasks. In a detailed error analysis we demonstrate that the distribution of this metric can be inspected by system developers to identify error types and system weaknesses.
Finally, we analyse the performance of two state-of-the-art \nlu{} systems on \asr{} data. 
We find that a sequential decoding approach for SLU, which starts from the more abstract notion of scenario and action produces better results for entity tagging, than an approach which works bottom up, i.e.\ starting from the entities. Our error analysis suggests that this is due to the former approach being able to better account for noise by priming entity tagging, which is a more challenging task than scenario or action recognition. 

In future work, we hope that \datasetacr{} will be a valuable resource for developing \etoe-\slu{} systems, as well as more traditional pipeline approaches to SLU.
The next step is to extend \datasetacr{} with spontaneous speech, which would again increase its complexity, but also move it one step closer to real-life applications.

\small
\section*{Acknowledgements}
Thanks to Emotech Ltd and H. Zhuang for agreeing to release this data for research purposes. Special thanks to P. Mediano, M. Zhou and X. Chen for help with designing and organising data collection.
This research received funding from the EPSRC project MaDrIgAL (EP/N017536/1), as well as Google Research Grant to support NLU and dialog research at Heriot-Watt University.
\normalsize

\bibliographystyle{acl_natbib}

\bibliography{main}

\begin{thebibliography}{38}
\expandafter\ifx\csname natexlab\endcsname\relax\def\natexlab#1{#1}\fi

\bibitem[{Ardila et~al.(2019)Ardila, Branson, Davis, Henretty, Kohler, Meyer,
  Morais, Saunders, Tyers, and Weber}]{ardila2019common}
Rosana Ardila, Megan Branson, Kelly Davis, Michael Henretty, Michael Kohler,
  Josh Meyer, Reuben Morais, Lindsay Saunders, Francis~M. Tyers, and Gregor
  Weber. 2019.
\newblock \href {http://arxiv.org/abs/1912.06670} {Common voice: A
  massively-multilingual speech corpus}.

\bibitem[{Bell et~al.(2020)Bell, Fainberg, Klejch, Li, Renals, and
  Swietojanski}]{bell2020adaptation}
Peter Bell, Joachim Fainberg, Ondrej Klejch, Jinyu Li, Steve Renals, and Pawel
  Swietojanski. 2020.
\newblock \href {http://arxiv.org/abs/2008.06580} {Adaptation algorithms for
  speech recognition: An overview}.

\bibitem[{Budzianowski et~al.(2018)Budzianowski, Wen, Tseng, Casanueva, Stefan,
  Osman, and Ga{\v{s}}i\'c}]{budzianowski2018:woz}
Pawe{\l} Budzianowski, Tsung-Hsien Wen, Bo-Hsiang Tseng, I{\~n}igo Casanueva,
  Ultes Stefan, Ramadan Osman, and Milica Ga{\v{s}}i\'c. 2018.
\newblock Multiwoz - a large-scale multi-domain wizard-of-oz dataset for
  task-oriented dialogue modelling.
\newblock In \emph{Proceedings of the 2018 Conference on Empirical Methods in
  Natural Language Processing (EMNLP)}.

\bibitem[{Carletta(2007)}]{carletta2007unleashing}
Jean Carletta. 2007.
\newblock Unleashing the killer corpus: experiences in creating the
  multi-everything ami meeting corpus.
\newblock \emph{Language Resources and Evaluation}, 41(2):181--190.

\bibitem[{Cieri et~al.(2004)Cieri, Miller, and Walker}]{cieri2004fisher}
Christopher Cieri, David Miller, and Kevin Walker. 2004.
\newblock The fisher corpus: a resource for the next generations of
  speech-to-text.
\newblock In \emph{LREC}, volume~4, pages 69--71.

\bibitem[{Coucke et~al.(2018)Coucke, Saade, Ball, Bluche, Caulier, Leroy,
  Doumouro, Gisselbrecht, Caltagirone, Lavril, Primet, and
  Dureau}]{Coucke18:Snips}
Alice Coucke, Alaa Saade, Adrien Ball, Th{\'e}odore Bluche, Alexandre Caulier,
  David Leroy, Cl{\'e}ment Doumouro, Thibault Gisselbrecht, Francesco
  Caltagirone, Thibaut Lavril, Ma{\"e}l Primet, and Joseph Dureau. 2018.
\newblock Snips voice platform: an embedded spoken language understanding
  system for private-by-design voice interfaces.
\newblock \emph{ArXiv}, abs/1805.10190.

\bibitem[{Covington et~al.(2006)Covington, He, Brown-Johnson, Naci, and
  Brown}]{covington06:dlevel}
Michael Covington, Congzhou He, Cati Brown-Johnson, Lorina Naci, and John
  Brown. 2006.
\newblock How complex is that sentence? a proposed revision of the rosenberg
  and abbeduto d-level scale.

\bibitem[{Dahl et~al.(1994)Dahl, Bates, Brown, Fisher, Hunicke-smith, Pallett,
  Rudnicky, and Shriberg}]{Dahl:94ATIS}
Deborah~A. Dahl, Madeleine Bates, Michael Brown, William Fisher, Kate
  Hunicke-smith, David Pallett, Er~Rudnicky, and Elizabeth Shriberg. 1994.
\newblock Expanding the scope of the {ATIS} task: the {ATIS-3} corpus.
\newblock In \emph{in Proc. ARPA Human Language Technology Workshop '92,
  Plainsboro, NJ}, pages 43--48. Morgan Kaufmann.

\bibitem[{{Desot} et~al.(2019){Desot}, {Portet}, and {Vacher}}]{desot:asru2019}
T.~{Desot}, F.~{Portet}, and M.~{Vacher}. 2019.
\newblock Slu for voice command in smart home: Comparison of pipeline and
  end-to-end approaches.
\newblock In \emph{2019 IEEE Automatic Speech Recognition and Understanding
  Workshop (ASRU)}, pages 822--829.

\bibitem[{E et~al.(2019)E, Niu, Chen, and Song}]{e-etal-2019-novel}
Haihong E, Peiqing Niu, Zhongfu Chen, and Meina Song. 2019.
\newblock \href {https://doi.org/10.18653/v1/P19-1544} {A novel bi-directional
  interrelated model for joint intent detection and slot filling}.
\newblock In \emph{Proceedings of the 57th Annual Meeting of the Association
  for Computational Linguistics}, pages 5467--5471, Florence, Italy.
  Association for Computational Linguistics.

\bibitem[{Godfrey et~al.(1992)Godfrey, Holliman, and
  McDaniel}]{Godfrey92:switchboard}
John~J. Godfrey, Edward~C. Holliman, and Jane McDaniel. 1992.
\newblock Switchboard: Telephone speech corpus for research and development.
\newblock In \emph{Proceedings of the 1992 IEEE International Conference on
  Acoustics, Speech and Signal Processing - Volume 1}, ICASSP’92, page
  517–520, USA. IEEE Computer Society.

\bibitem[{Haghani et~al.(2018)Haghani, Narayanan, Bacchiani, Chuang, Gaur,
  Moreno, Prabhavalkar, Qu, and Waters}]{Haghani2018:FromAT}
Parisa Haghani, Arun Narayanan, Michiel Bacchiani, Galen Chuang, Neeraj Gaur,
  Pedro~J. Moreno, Rohit Prabhavalkar, Zhongdi Qu, and Austin Waters. 2018.
\newblock From audio to semantics: Approaches to end-to-end spoken language
  understanding.
\newblock \emph{2018 IEEE Spoken Language Technology Workshop (SLT)}, pages
  720--726.

\bibitem[{Hemphill et~al.(1990)Hemphill, Godfrey, and
  Doddington}]{Hemphill90:ATIS}
Charles~T. Hemphill, John~J. Godfrey, and George~R. Doddington. 1990.
\newblock \href {https://doi.org/10.3115/116580.116613} {The atis spoken
  language systems pilot corpus}.
\newblock In \emph{Proceedings of the Workshop on Speech and Natural Language},
  HLT ’90, page 96–101, USA. Association for Computational Linguistics.

\bibitem[{Janin et~al.(2003)Janin, Baron, Edwards, Ellis, Gelbart, Morgan,
  Peskin, Pfau, Shriberg, Stolcke et~al.}]{janin2003icsi}
Adam Janin, Don Baron, Jane Edwards, Dan Ellis, David Gelbart, Nelson Morgan,
  Barbara Peskin, Thilo Pfau, Elizabeth Shriberg, Andreas Stolcke, et~al. 2003.
\newblock The icsi meeting corpus.
\newblock In \emph{2003 IEEE International Conference on Acoustics, Speech, and
  Signal Processing, 2003. Proceedings.(ICASSP'03).}, volume~1, pages I--I.
  IEEE.

\bibitem[{Johnson(1944)}]{wendell44:lex}
Wendell Johnson. 1944.
\newblock \href
  {https://search.alexanderstreet.com/view/work/bibliographic_entity%7Cbibliographic_details%7C2140269}
  {\emph{Studies in Language Behavior: I. A Program of Research}}.
\newblock National Foreign Language Center Technical Reports. American
  Psychological Association, Psychological Monographs: General and Applied.

\bibitem[{Jurafsky and Shriberg(1997)}]{Jurafsky1997:DASwitchboard}
Dan Jurafsky and Elizabeth Shriberg. 1997.
\newblock Switchboard-damsl labeling project coder''s manual.

\bibitem[{Ko et~al.(2017)Ko, Peddinti, Povey, Seltzer, and
  Khudanpur}]{ko2017study}
Tom Ko, Vijayaditya Peddinti, Daniel Povey, Michael~L Seltzer, and Sanjeev
  Khudanpur. 2017.
\newblock A study on data augmentation of reverberant speech for robust speech
  recognition.
\newblock In \emph{2017 IEEE International Conference on Acoustics, Speech and
  Signal Processing (ICASSP)}, pages 5220--5224. IEEE.

\bibitem[{Laufer(1994)}]{laufer94:lex}
Batia Laufer. 1994.
\newblock \href {https://doi.org/10.1177/003368829402500202} {The lexical
  profile of second language writing: Does it change over time?}
\newblock \emph{RELC Journal}, 25(2):21--33.

\bibitem[{Liu et~al.(2019)Liu, Eshghi, Swietojanski, and
  Rieser}]{liu19:benchmarking}
Xingkun Liu, Arash Eshghi, Pawel Swietojanski, and Verena Rieser. 2019.
\newblock Benchmarking natural language understanding services for building
  conversational agents.
\newblock In \emph{10th International Workshop on Spoken Dialogue Systems
  Technology}.

\bibitem[{Lu(2009)}]{lu09:synt}
Xiaofei Lu. 2009.
\newblock Automatic measurement of syntactic complexity in child language
  acquisition.

\bibitem[{Lu(2012)}]{lu2012:lex}
Xiaofei Lu. 2012.
\newblock \href
  {https://onlinelibrary.wiley.com/doi/abs/10.1111/j.1540-4781.2011.01232_1.x}
  {The relationship of lexical richness to the quality of esl learners’ oral
  narratives}.
\newblock \emph{The Modern Language Journal}, 96(2):190--208.

\bibitem[{Lugosch et~al.(2019{\natexlab{a}})Lugosch, Meyer, Nowrouzezahrai, and
  Ravanelli}]{arxiv.org/abs/1910.09463}
Loren Lugosch, Brett Meyer, Derek Nowrouzezahrai, and Mirco Ravanelli.
  2019{\natexlab{a}}.
\newblock \href {http://arxiv.org/abs/1910.09463} {Using speech synthesis to
  train end-to-end spoken language understanding models}.
\newblock abs/1910.09463.

\bibitem[{Lugosch et~al.(2019{\natexlab{b}})Lugosch, Ravanelli, Ignoto, Tomar,
  and Bengio}]{lugosh19:Interspeech}
Loren Lugosch, Mirco Ravanelli, Patrick Ignoto, Vikrant~Singh Tomar, and Yoshua
  Bengio. 2019{\natexlab{b}}.
\newblock \href {https://doi.org/10.21437/Interspeech.2019-2396} {Speech model
  pre-training for end-to-end spoken language understanding}.
\newblock In \emph{Interspeech 2019, 20th Annual Conference of the
  International Speech Communication Association, Graz, Austria, 15-19
  September 2019}, pages 814--818. {ISCA}.

\bibitem[{Marino and Hain(2011)}]{Marino2011AnAO}
Davide Marino and Thomas Hain. 2011.
\newblock An analysis of automatic speech recognition with multiple
  microphones.
\newblock In \emph{INTERSPEECH}.

\bibitem[{Miksik et~al.(2020)Miksik, Munasinghe, Asensio-Cubero, Bethi, Huang,
  Zylfo, Liu, Nica, Mitrocsak, Mezza, Beard, Shi, Ng, Mediano, Fountas, Lee,
  Medvesek, Zhuang, Rogers, and Swietojanski}]{miksik2020building}
O.~Miksik, I.~Munasinghe, J.~Asensio-Cubero, S.~Reddy Bethi, S-T. Huang,
  S.~Zylfo, X.~Liu, T.~Nica, A.~Mitrocsak, S.~Mezza, R.~Beard, R.~Shi, R.~Ng,
  P.~Mediano, Z.~Fountas, S-H. Lee, J.~Medvesek, H.~Zhuang, Y.~Rogers, and
  P.~Swietojanski. 2020.
\newblock \href {http://arxiv.org/abs/2005.01322} {Building proactive voice
  assistants: When and how (not) to interact}.
\newblock \emph{CoRR}, abs/2005.01322.

\bibitem[{Novikova et~al.(2017)Novikova, Du{\v{s}}ek, and
  Rieser}]{novikova-etal-2017-e2e}
Jekaterina Novikova, Ond{\v{r}}ej Du{\v{s}}ek, and Verena Rieser. 2017.
\newblock \href {https://doi.org/10.18653/v1/W17-5525} {The {E}2{E} dataset:
  New challenges for end-to-end generation}.
\newblock In \emph{Proceedings of the 18th Annual {SIG}dial Meeting on
  Discourse and Dialogue}, pages 201--206, Saarbr{\"u}cken, Germany.
  Association for Computational Linguistics.

\bibitem[{Panayotov et~al.(2015)Panayotov, Chen, Povey, and
  Khudanpur}]{panayotov2015librispeech}
Vassil Panayotov, Guoguo Chen, Daniel Povey, and Sanjeev Khudanpur. 2015.
\newblock Librispeech: an asr corpus based on public domain audio books.
\newblock In \emph{2015 IEEE International Conference on Acoustics, Speech and
  Signal Processing (ICASSP)}, pages 5206--5210. IEEE.

\bibitem[{Peddinti et~al.(2015)Peddinti, Povey, and
  Khudanpur}]{peddinti2015time}
Vijayaditya Peddinti, Daniel Povey, and Sanjeev Khudanpur. 2015.
\newblock A time delay neural network architecture for efficient modeling of
  long temporal contexts.
\newblock In \emph{Sixteenth Annual Conference of the International Speech
  Communication Association}.

\bibitem[{Povey et~al.(2011)Povey, Ghoshal, Boulianne, Burget, Glembek, Goel,
  Hannemann, Motlicek, Qian, Schwarz et~al.}]{povey2011kaldi}
Daniel Povey, Arnab Ghoshal, Gilles Boulianne, Lukas Burget, Ondrej Glembek,
  Nagendra Goel, Mirko Hannemann, Petr Motlicek, Yanmin Qian, Petr Schwarz,
  et~al. 2011.
\newblock The kaldi speech recognition toolkit.
\newblock In \emph{IEEE 2011 workshop on automatic speech recognition and
  understanding}, CONF. IEEE Signal Processing Society.

\bibitem[{Povey et~al.(2016)Povey, Peddinti, Galvez, Ghahremani, Manohar, Na,
  Wang, and Khudanpur}]{povey2016purely}
Daniel Povey, Vijayaditya Peddinti, Daniel Galvez, Pegah Ghahremani, Vimal
  Manohar, Xingyu Na, Yiming Wang, and Sanjeev Khudanpur. 2016.
\newblock Purely sequence-trained neural networks for asr based on lattice-free
  mmi.

\bibitem[{Rastogi et~al.(2020)Rastogi, Zang, Sunkara, Gupta, and
  Khaitan}]{48919}
Abhinav Rastogi, Xiaoxue Zang, Srinivas~Kumar Sunkara, Raghav Gupta, and Pranav
  Khaitan. 2020.
\newblock Schema-guided dialogue state tracking task at dstc8.
\newblock In \emph{AAAI Dialog System Technology Challenges Workshop}.

\bibitem[{Schuster et~al.(2019)Schuster, Gupta, Shah, and
  Lewis}]{Schuster19:facebook}
Sebastian Schuster, Sonal Gupta, Rushin Shah, and Mike Lewis. 2019.
\newblock \href {https://doi.org/10.18653/v1/n19-1380} {Cross-lingual transfer
  learning for multilingual task oriented dialog}.
\newblock In \emph{Proceedings of the 2019 Conference of the North American
  Chapter of the Association for Computational Linguistics: Human Language
  Technologies, {NAACL-HLT} 2019, Minneapolis, MN, USA, June 2-7, 2019, Volume
  1 (Long and Short Papers)}, pages 3795--3805. Association for Computational
  Linguistics.

\bibitem[{Serdyuk et~al.(2018)Serdyuk, Wang, Fuegen, Kumar, Liu, and
  Bengio}]{Serdyuk18:towards}
Dmitriy Serdyuk, Yongqiang Wang, Christian Fuegen, Anuj Kumar, Baiyang Liu, and
  Yoshua Bengio. 2018.
\newblock \href {http://arxiv.org/abs/1802.08395} {Towards end-to-end spoken
  language understanding}.
\newblock \emph{CoRR}, abs/1802.08395.

\bibitem[{Swietojanski et~al.(2016)Swietojanski, Li, and
  Renals}]{swietojanski2016learning}
Pawel Swietojanski, Jinyu Li, and Steve Renals. 2016.
\newblock Learning hidden unit contributions for unsupervised acoustic model
  adaptation.
\newblock \emph{IEEE/ACM Transactions on Audio, Speech, and Language
  Processing}, 24(8):1450--1463.

\bibitem[{Tomashenko et~al.(2019)Tomashenko, Caubri{\`e}re, and
  Est{\`e}ve}]{tomashenko:hal-02307811}
Natalia Tomashenko, Antoine Caubri{\`e}re, and Yannick Est{\`e}ve. 2019.
\newblock \href {https://doi.org/10.21437/Interspeech.2019-2158}
  {{Investigating Adaptation and Transfer Learning for End-to-End Spoken
  Language Understanding from Speech}}.
\newblock In \emph{{Interspeech 2019}}, pages 824--828, Graz, Austria. {ISCA}.

\bibitem[{Vanzo et~al.(2019)Vanzo, Bastianelli, and Lemon}]{vanzo:2019b}
Vanzo, Bastianelli, and Lemon. 2019.
\newblock \href {https://doi.org/10.18653/v1/W19-5931} {Hierarchical multi-task
  natural language understanding for cross-domain conversational {AI}: {HERMIT}
  {NLU}}.
\newblock In \emph{Proceedings of the 20th Annual SIGdial Meeting on Discourse
  and Dialogue}, pages 254--263, Stockholm, Sweden. Association for
  Computational Linguistics.

\bibitem[{Wolfe-Quintero et~al.(1998)Wolfe-Quintero, Inagaki, and
  Kim}]{wolfe1998:second}
K.~Wolfe-Quintero, S.~Inagaki, and H.Y. Kim. 1998.
\newblock \href {https://books.google.nl/books?id=IboEPPjPGgkC} {\emph{Second
  Language Development in Writing: Measures of Fluency, Accuracy, \&
  Complexity}}.
\newblock National Foreign Language Center Technical Reports. Second Language
  Teaching \& Curriculum Center, University of Hawaii at Manoa.

\bibitem[{Zhu et~al.(2019)Zhu, Zhao, Zhao, Zong, and
  Yu}]{10.1145/3340555.3356098}
Su~Zhu, Zijian Zhao, Tiejun Zhao, Chengqing Zong, and Kai Yu. 2019.
\newblock \href {https://doi.org/10.1145/3340555.3356098} {Catslu: The 1st
  chinese audio-textual spoken language understanding challenge}.
\newblock In \emph{2019 International Conference on Multimodal Interaction},
  ICMI '19, page 521–525, New York, NY, USA. Association for Computing
  Machinery.

\end{thebibliography}


\end{document}